\pdfoutput=1

\documentclass[11pt]{article}

\usepackage[final]{acl}

\usepackage{times}
\usepackage{latexsym}
\usepackage[T1]{fontenc}
\usepackage[utf8]{inputenc}
\usepackage{microtype}
\usepackage{inconsolata}
\usepackage{graphicx}
\usepackage{amsmath}
\usepackage{booktabs}
\usepackage{subcaption}
\usepackage{enumitem}
\usepackage{float}
\usepackage[table]{xcolor}
\usepackage[utf8]{inputenc}
\usepackage[T1]{fontenc}
\usepackage{setspace}
\usepackage{enumitem}
\DeclareUnicodeCharacter{202F}{\,}
\DeclareUnicodeCharacter{060C}{,}
\title{SearchInstruct: Enhancing Domain Adaptation via Retrieval-Based Instruction Dataset Creation}

\author{
  Iman Barati\textsuperscript{1} \quad
  Mostafa Amiri\textsuperscript{2} \quad
  Heshaam Faili\textsuperscript{2} \\[0.3em] 
  \textsuperscript{1} Iran University of Science and Technology \quad
  \textsuperscript{2} University of Tehran \\[0.2em]
  \texttt{iman\_barati@comp.iust.ac.ir} \quad
  \texttt{\{mostafa.amiri,hfaili\}@ut.ac.ir}
}

\begin{document}
\maketitle

\begin{abstract}
Supervised Fine-Tuning (SFT) is essential for training large language models (LLMs), significantly enhancing critical capabilities such as instruction-following and in-context learning. Nevertheless, creating suitable training datasets tailored for specific domains remains challenging due to unique domain constraints and data scarcity.

In this paper, we propose \textit{SearchInstruct}, an innovative method explicitly designed to construct high-quality instruction datasets for SFT. Our approach begins with a limited set of domain-specific, human-generated questions, which are systematically expanded using a large language model. Subsequently, domain-relevant resources are dynamically retrieved to generate accurate and contextually appropriate answers for each augmented question.

Experimental evaluation demonstrates that \textit{SearchInstruct} notably enhances both the diversity and quality of SFT datasets, leading to measurable improvements in LLM performance within specialized domains. Additionally, we demonstrate that beyond dataset generation, the proposed method can also effectively facilitate tasks such as model editing, enabling efficient updates to existing models.

To facilitate reproducibility and community adoption, we provide full implementation details, the complete set of generated instruction–response pairs, and the source code in a publicly accessible Git repository.\footnote{\url{https://github.com/mostafaamiri/SearchInstruct}}
\end{abstract}

\section{Introduction}

The rapid progress of LLMs and their success in solving general-purpose natural language processing (NLP) tasks~\citep{wei2021finetuned, sanh2022multitask} have led to a growing focus on adapting these models to specialized domains. Achieving this goal requires not only access to cleaned raw corpora, but also the construction of structured datasets suitable for the SFT stage~\citep{ouyang2022training}.

Generating high-quality SFT data—especially for expert domains—presents several major challenges. The data must be both diverse and capable of reflecting the inherent complexity of real-world user queries. Existing approaches either rely on extensive manual annotation, as in InstructGPT~\citep{ouyang2022training}, suffer from limited coverage and task diversity~\citep{ziegler2024craft}, or fail to simulate real user needs accurately. Furthermore, many methods depend solely on the internal knowledge of LLMs, which may be outdated or insufficient for domain-specific adaptation~\citep{lewis2020retrieval,lazaridou2022internet}.

At the same time, LLMs have shown that—when provided with appropriate context—they can generate accurate and informative responses to even complex domain-specific questions~\citep{izacard2022atlas}. This observation suggests that, by systematically generating diverse and realistic queries and retrieving relevant, up-to-date contextual information, it is possible to construct powerful training datasets without relying on human annotation. Such an approach enables the integration of external domain-specific knowledge into the training process, offering a path to updating and extending model capabilities~\citep{guu2020realm, wang2023instructretro}.

In this paper, we introduce \textit{SearchInstruct}, a novel framework for automatically generating supervised training data. The method operates by expanding a small set of seed questions using an LLM, then dynamically retrieving relevant documents to construct accurate, context-grounded answers. Our empirical results demonstrate that this approach not only increases the quality and diversity of SFT data, but also effectively enhances and updates LLM behavior in specialized domains by leveraging up-to-date external knowledge.

Our contributions are as follows:
\begin{enumerate}[itemsep=0.1em, topsep=0.3em]
    \item We introduce \textit{SearchInstruct}, an automated framework for constructing SFT datasets by combining targeted document retrieval with grounded answer generation;
    \item We design and implement a four-stage pipeline for query expansion, diversification, document retrieval, and answer synthesis;
    \item We empirically show that, in certain specialized domains, this pipeline leads to improvements in model performance and instruction-following accuracy;
    \item We demonstrate the effectiveness of \textit{SearchInstruct} in model editing scenarios, enabling efficient updates to model outputs using newly acquired knowledge.
\end{enumerate}

\section{Related Work}

Instruction tuning has emerged as a key strategy for adapting LLMs to follow user instructions. The FLAN approach by Wei et al.~\citep{wei2021finetuned} demonstrated that fine-tuning on a broad range of instruction-formatted tasks leads to strong zero-shot generalization. Similarly, T0 by Sanh et al.~\citep{sanh2022multitask} employed multitask prompted training to enable cross-task generalization.

Wang et al.~\citep{wang2022supernaturalinstructions} extended this paradigm with the Super-NaturalInstructions dataset, collecting over 1,600 crowdsourced tasks paired with declarative instructions to promote task diversity. InstructGPT~\citep{ouyang2022training} introduced a method for aligning LLMs with human preferences using instruction–response pairs ranked via human feedback. Despite their utility, these datasets are static, expensive to construct, and often insufficient for domain-specific applications or ongoing model updates.

To address limitations in scalability and recency, Self-Instruct~\citep{wang2022selfinstruct} bootstraps a base LLM to generate new instructions and responses from a small seed set, filtering outputs using heuristics. While it enhances general instruction-following capabilities, it still relies entirely on the model’s internal knowledge, which may be outdated or incomplete for specialized domains.

Recent work has focused on automating instruction dataset creation to reduce dependence on manual annotation. The Alpaca project~\citep{taori2023alpaca} used GPT-3.5 to generate 52,000 instruction–response pairs, enabling effective fine-tuning of a LLaMA model. Evol-Instruct by Xu et al.~\citep{xu2023wizardlm} introduced an evolutionary strategy to rewrite and complicate instructions, leading to the creation of WizardLM. InstructZero~\citep{chen2023instructzero} proposed a label-free tuning method using reinforcement learning with black-box models. While these approaches improve scale and diversity, they generally lack grounding in real-world evidence and do not facilitate domain-specific adaptation.

Retrieval-augmented methods aim to improve factuality and relevance by incorporating external documents. RAG~\citep{lewis2020retrieval} combines dense retrieval with generation to condition responses on retrieved passages. REALM~\citep{guu2020realm} pre-trains language models with latent document retrieval to support open-domain QA. More recently, InstructRetro~\citep{wang2023instructretro} integrates retrieval with instruction tuning to enhance grounding, and RA-DIT~\citep{lin2023radit} jointly trains a retriever and generator using dual instruction signals. However, these methods often require costly retriever training and may not be ideal for rapid, low-resource domain updates.

\textit{SearchInstruct} addresses the limitations of static and hallucinated instruction datasets by incorporating real-time document retrieval during data generation. Unlike Self-Instruct, which depends solely on internal knowledge, our method grounds both the instruction and the response in up-to-date, domain-specific documents. Unlike methods such as RAG and REALM, which apply retrieval at inference time, SearchInstruct applies it during dataset construction—allowing the fine-tuned model to internalize retrieved knowledge. This enables scalable, lightweight domain adaptation and continual knowledge refresh, bridging the gap between static SFT and real-world evolving needs.

\section{Methodology}

The core idea behind our method is that an LLM, when provided with relevant contextual information—even for domain-specific queries—can generate accurate and informative answers. These instruction–response pairs can then be used as high-quality data for SFT, improving the model’s ability to follow instructions and generalize in specialized domains.

\begin{figure*}[t]
    \centering
    \includegraphics[width=0.9\textwidth]{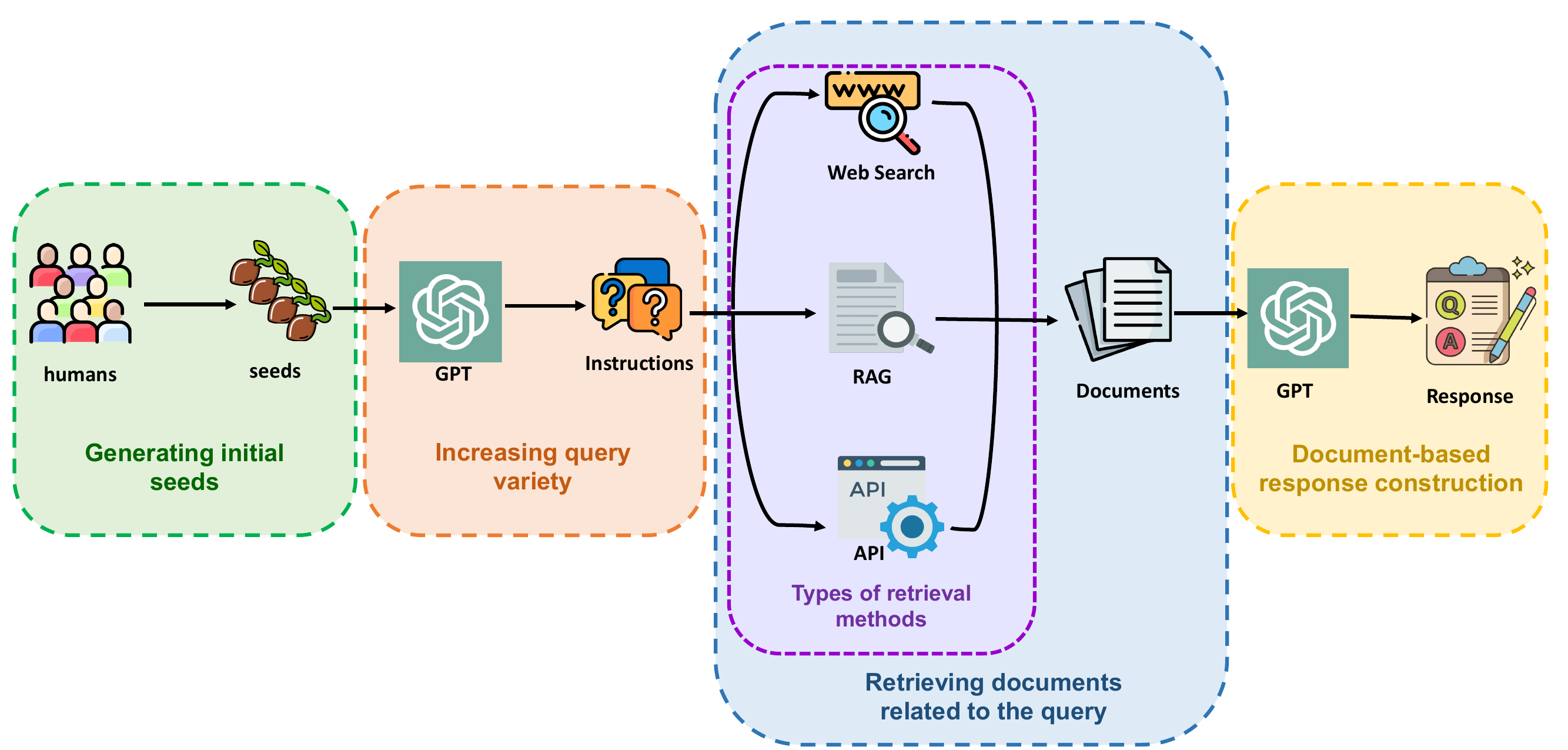}
    \caption{The four-stage \textit{SearchInstruct} pipeline: seed generation, query expansion, document retrieval, and response construction.}
    \label{fig:pipeline}
\end{figure*}

While prior work such as Self-Instruct~\citep{wang2022selfinstruct} demonstrated the value of generating large quantities of instruction–response examples, we argue that in the current landscape—where LLMs often outperform crowd-annotated answers—the primary focus should shift from generating more answers to diversifying the questions themselves. A broader and more realistic query distribution enables better coverage of domain-specific tasks, especially when answers are grounded in retrieved, up-to-date evidence.

To operationalize this idea, we propose a four-stage pipeline, illustrated in Figure~\ref{fig:pipeline}, consisting of the following steps:
\begin{enumerate}[itemsep=0.1em, topsep=0.3em]
    \item \textbf{Seed Generation}: Start with a small set of domain-relevant, human-written or curated instruction prompts;
    \item \textbf{Query Expansion}: Use an LLM to generate diverse variations and paraphrases of the seed prompts;
    \item \textbf{Document Retrieval}: Retrieve domain-specific and up-to-date documents relevant to each expanded query (using web search, RAG, or external APIs);
    \item \textbf{Response Construction}: Generate accurate, context-aware answers grounded in the retrieved evidence.
\end{enumerate}

In the following sections, we describe each stage of the pipeline in detail.

\subsection{\textit{SearchInstruct} Dataset Formulation}

We formally define our instruction dataset $\mathcal{D}$ as a set of instruction–response pairs constructed through a four-stage pipeline.

Let $Q = \{q_1, q_2, \dots, q_{|Q|}\}$ denote the initial set of domain-specific seed queries. From $Q$, we generate a set of additional instructions via LLM-based expansion. Let $I$ denote the set of newly generated instructions, and define the final instruction pool as:
\[
I^{\text{total}} = Q \cup I
\]

For each instruction $i \in I^{\text{total}}$, we retrieve a set of relevant textual contexts (e.g., documents, passages), denoted $C_i = \{c_{i,1}, \dots, c_{i,k}\}$. We then use an LLM to generate an answer $A_i$ grounded in the retrieved context:
\[
A_i = \mathrm{LLM}_{\mathrm{answer}}(i, C_i^{\text{filtered}})
\]

The final instruction dataset is constructed as:
\[
\mathcal{D} = \left\{(i, A_i) \;\middle|\; i \in I^{\text{total}} \right\}
\]

During training, the model $M$ learns the conditional instruction-following mapping:
\[
M(i) \rightarrow A_i
\]

Note that while $C_i$ is essential for answer generation during dataset construction, it is not required at inference or training time. This enables $M$ to internalize external knowledge through SFT and improve response quality in dynamic or specialized domains.

\subsection{Stage 1: Seed Generation}
\label{sec:seed-generation}

The \textit{SearchInstruct} pipeline begins with the creation of a diverse set of seed queries $Q = \{q_1, q_2, \dots, q_{|Q|}\}$ that serve as the foundation for subsequent query expansion. Because all downstream instructions are generated based on these seeds, ensuring high quality and topical diversity in $Q$ is critical.

We consider two approaches for seed construction:

\paragraph{Human-Crafted Seeds.}  
In the fully manual setup, domain experts are given a detailed instruction-writing guideline that includes:

\begin{itemize}[itemsep=0.1em, topsep=0.3em, leftmargin=1.5em]
    \item Requirements for seed diversity and realism,
    \item Types of questions that are difficult to synthesize from retrieved documents (e.g., abstract reasoning, subjective judgment),
    \item Subcategories within the target domain.
\end{itemize}
Each annotator is asked to write several domain-specific instructions based on this guideline. This ensures that $Q$ contains high-coverage, realistic questions that may not emerge from automatic generation methods.

\paragraph{Human–LLM Collaborative Seeds.}  
In the hybrid setup, annotators are guided to use powerful LLMs (e.g., GPT-4o, Claude 3.5 Sonnet, Gemini 2 Pro) to generate seeds. A separate guideline is provided, which instructs annotators to:

\begin{enumerate}[itemsep=0.1em, topsep=0.3em]
    \item \setstretch{0.9}Ask the LLM to suggest a comprehensive list of subtopics within the target domain;
    \item \setstretch{0.9}For each subtopic, prompt the LLM to generate multiple instruction-style queries based on predefined types;
    \item \setstretch{0.9}Select, edit, or rewrite these outputs into final seeds.
\end{enumerate}

This method enables rapid scaling while retaining human supervision and correction. In practice, we collected approximately twice as many seeds per annotator compared to the fully manual setting.
  
After initial training, we identify weaknesses in specific subdomains or instruction types. Additional seeds are then generated—using either method above—to target these weaknesses and enrich low-performing areas.

\hyperref[subsec:seed-generation]{Appendix~C} provides a detailed example of how seed queries were constructed in the domain of Iranian culinary and tourism.

\subsection{Stage 2: Query Expansion}

To increase the diversity and coverage of the instruction dataset, we expand the initial seed set $Q$ using a LLM. This process is inspired by the Self-Instruct framework~\citep{wang2022selfinstruct}, but adapted to focus exclusively on instruction generation rather than full instruction–response pairs.

At each expansion step, a subset of $n$ seed queries $S = \{q_1, q_2, \dots, q_n\} \subset Q$ is selected, and the LLM is prompted to synthesize $k$ new instruction-style queries:
\[
I_S = \{i_1, i_2, \dots, i_k\}
\]
Each instruction $i_j \in I_S$ is either a paraphrase of a single $q_i$ or a novel combination of multiple seed queries in $S$. This process is repeated over $m$ iterations to generate a sufficiently diverse instruction pool.

We denote the union of all generated instructions across iterations as:
\[
I = \bigcup_{t=1}^{m} I_{S_t}
\]
The final instruction set used for downstream processing is defined as:
\[
I^{\text{total}} = Q \cup I
\]
\hyperref[subsec:query-expansion]{Appendix A.1} shows the prompt template used to diversify queries during expansion.

\subsection{Stage 3: Document Retrieval}

In this stage, we retrieve domain-relevant content to ground each instruction $i \in I^{\text{total}}$ in external evidence. The goal is to identify high-quality sources that can support accurate and informative answer generation in the next stage.

Depending on the application setting and available resources, we consider multiple retrieval strategies:

\begin{itemize}
    \item \textbf{RAG-style retrieval:} If access to a structured document collection is available, we apply dense or hybrid retrieval to identify the top-$k$ text chunks $C_i = \{c_{i,1}, \dots, c_{i,k}\}$ that are semantically similar to the instruction $i$. This is done using retriever models such as DPR or OpenAI embeddings.
    
    \item \textbf{Web search:} For open-domain instructions or in the absence of curated corpora, we use web search engines to collect a set of relevant URLs and associated snippets as external context.
    
    \item \textbf{External APIs:} In some cases, domain-specific APIs (e.g., tourism databases, legal knowledge graphs) are queried to extract structured or semi-structured evidence.
\end{itemize}

To improve the effectiveness of retrieval, we do not use the instruction $i$ directly. Instead, we construct a search-oriented query $q_i^{\text{search}}$ from $i$ using an LLM-based rewriting module:
\[
q_i^{\text{search}} = \mathrm{LLM}_{\mathrm{rewriter}}(i)
\]

This transformation helps reduce ambiguity and match the query to the structure and vocabulary of the underlying retrieval source.

Prompt template used for query rewriting is provided in \hyperref[appendix:search_prompt]{Appendix~A.2}.

\subsection{Stage 4: Response Construction}

In the final stage of the \textit{SearchInstruct} pipeline, we synthesize high-quality answers for each instruction $i \in I^{\text{total}}$ using retrieved evidence. Each instruction is paired with its corresponding set of retrieved contexts $C_i = \{c_{i,1}, \dots, c_{i,k}\}$ obtained from Stage~3.

To ensure efficient and accurate generation, we pre-process the retrieved documents to filter out irrelevant or noisy content. This step is especially important when dealing with long documents, as excessive input length can degrade LLM performance and increase computational cost. We apply one of the following filtering strategies:

\begin{itemize}
    \item \textbf{LLM-based chunk filtering:} A lightweight language model is used to rank or extract segments of each document that are most relevant to the instruction;
    \item \textbf{Rule-based filtering:} Heuristics are applied to remove common noise sources such as HTML tags, ads, metadata, or user comments.
\end{itemize}

After filtering, we concatenate the instruction $i$ with the cleaned context $C_i^{\text{filtered}}$ and prompt a powerful LLM to generate a contextually grounded answer:
\[
A_i = \mathrm{LLM}_{\mathrm{answer}}(i, C_i^{\text{filtered}})
\]

This produces the final instruction–response pair $(i, A_i)$ to be included in the training dataset $\mathcal{D}$.

Prompt templates for this stage are provided in \hyperref[appendix:response]{Appendix A.3}.

\section{Applications of \textit{SearchInstruct}}

The advantages of our proposed approach can be summarized in three key aspects:

\begin{enumerate}
    \item It enables the generation of accurate and contextually grounded responses for questions that require domain-specific knowledge—particularly in cases where no existing LLM is capable of providing a correct answer.

    \item The instruction–response pairs generated by this method provide a close approximation of real end-user queries.

    \item The method remains effective even when using smaller or open-source LLMs. As long as relevant documents are retrieved and appropriately incorporated, it is possible to construct high-quality responses to complex queries without relying on large proprietary models.
\end{enumerate}

In this work, we apply \textit{SearchInstruct} in two primary settings:

\begin{itemize}
    \item Constructing SFT datasets in the domain of Iranian culture, with a focus on two subdomains: traditional cuisine and domestic tourism;

    \item Updating model knowledge with up-to-date information, to enhance its ability to respond to recent or evolving queries in specialized domains.
\end{itemize}

In the following sections, we provide a detailed analysis of both application scenarios.

\subsection{SFT Dataset Construction for Iranian Culture}

Iran is a vast and culturally diverse country, home to a wide range of historical sites, regional customs, and intangible heritage. Many of these cultural and geographical details are either underrepresented or entirely absent in current LLMs. To support this claim, \hyperref[appendix:b]{Appendix~B} presents examples of domain-specific queries for which existing models failed to provide correct answers, while the \textit{SearchInstruct} framework successfully generated accurate responses.

\begin{figure*}[t]
    \centering
    \begin{subfigure}{0.48\textwidth}
        \centering
        \includegraphics[width=\linewidth]{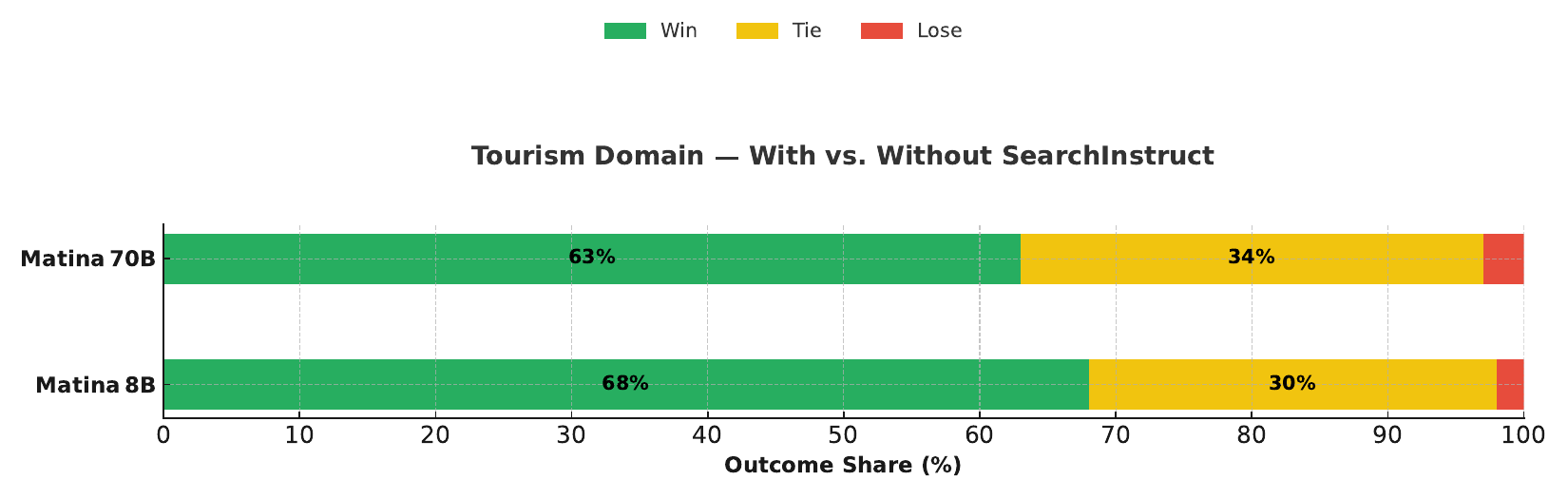}
        \caption{Tourism domain}
        \label{fig:tourism_chart}
    \end{subfigure}
    \hfill
    \begin{subfigure}{0.48\textwidth}
        \centering
        \includegraphics[width=\linewidth]{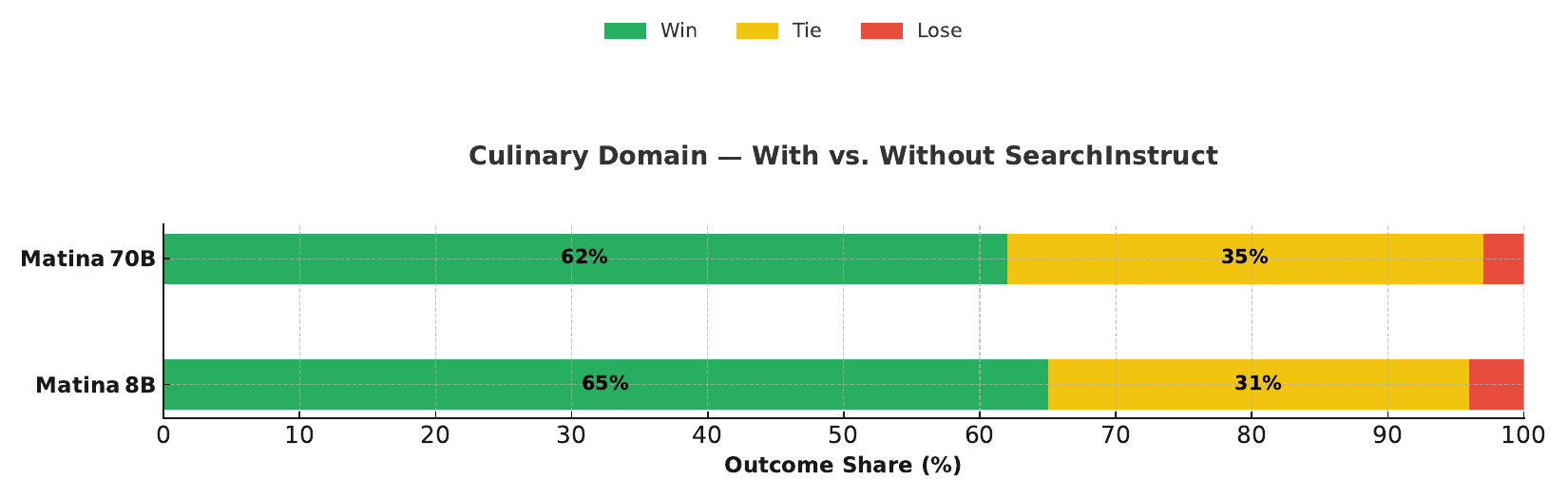}
        \caption{Culinary domain}
        \label{fig:culinary_chart}
    \end{subfigure}
    \caption{Human evaluation results comparing baseline models with those fine-tuned using the \textit{SearchInstruct} framework across two domains.}
    \label{fig:human_eval}
\end{figure*}

Another challenge is that Persian speakers, particularly in informal contexts, often use colloquial and context-rich language that differs significantly from formal documents. Except for a few major companies that serve Persian-speaking users, there is limited access to large-scale, real-world Persian queries. Attempting to synthesize such instructions purely from documents using LLMs often fails to capture the diversity of real user intent and task formulations. For example, a typical user might describe their own travel constraints and ask for a multi-day itinerary for a specific region of Iran—a type of instruction that is rarely (if ever) found explicitly in documents or websites.

To evaluate the practical effectiveness of our method, we applied \textit{SearchInstruct} in two cultural subdomains: traditional Iranian cuisine and domestic tourism. As discussed in the following sections, our results demonstrate the value of this approach in generating realistic, high-quality SFT data in resource-limited domains.

\subsubsection{Improvements in Cuisine and Tourism}

In the MATINA project ~\citep{hosseinbeigi2025matina}, multiple LoRA-tuned LLaMA-based models were trained across various domains, including traditional cuisine and tourism. The training data was generated using a mix of document-based question–answering and Evol-Instruct methods.

However, during human evaluation, it became evident that the models failed to perform adequately in several cultural subdomains. Upon analysis, these shortcomings were largely attributed to gaps in training data—particularly the absence of certain types of queries. These included questions such as:
\begin{itemize}[noitemsep, topsep=0pt]
    \item ``List multiple examples of...’’
    \item ``Imagine the following scenario...’’
    \item ``Recommend something based on personal constraints...’’
\end{itemize}
Such instructions are rarely present in documents and are not well-covered by typical synthetic data pipelines.

To address these gaps, we curated a specialized seed set as described in \hyperref[sec:seed-generation]{Section~\ref*{sec:seed-generation}} and \hyperref[subsec:seed-generation]{Appendix C}. The SearchInstruct pipeline was then applied to produce high-quality SFT data specifically targeting underrepresented question types and subdomains. The number of generated instances for each category using this method is reported in Table~\ref{tab:data_turism_culinary}. Each stage in the table corresponds to one full iteration of the SearchInstruct pipeline—including query expansion, document retrieval, and answer generation—which was repeated in response to human feedback or model performance analysis. In each stage, the pipeline was refined to specifically address the issues and gaps identified during the previous round, leading to incremental improvements in coverage, quality, and diversity.

\begin{figure*}[t]
  \centering
  \includegraphics[width=0.6\textwidth]{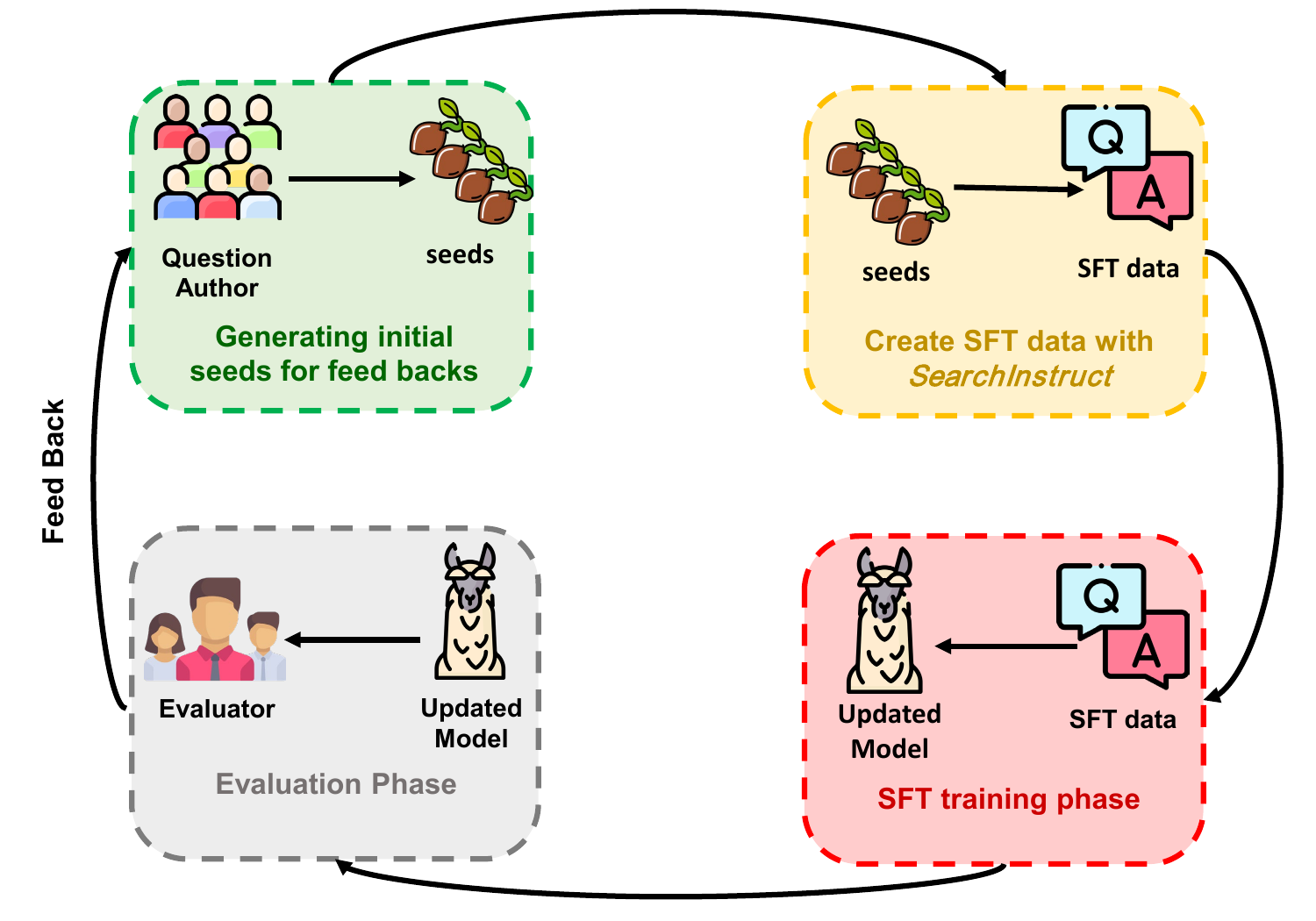}
  \caption{Iterative refinement loop enabled by the \textit{SearchInstruct} framework. After initial fine-tuning, specific model weaknesses are identified through targeted evaluation. New instruction–response data is then generated to directly address these shortcomings, creating a feedback loop that leads to focused and incremental improvements in model performance.}
  \label{fig:refinement_loop}
\end{figure*}

\begin{table}[h]
\centering
\renewcommand{\arraystretch}{1.25}  

\resizebox{0.9\linewidth}{!}{
\begin{tabular}{lrrr>{\columncolor{darkblue!10}\bfseries}r}
\toprule
\textbf{Domain} & \textbf{Stage 1} & \textbf{Stage 2} & \textbf{Stage 3} & \textbf{Overall} \\
\midrule
Culinary        & 4273             & 2886             & 1773             & 8932             \\
Tourism         & 3932             & 2378             & 1250             & 7560             \\
\bottomrule
\end{tabular}
}

\vspace{0.4em}  
\captionsetup{justification=raggedright,singlelinecheck=false}  
\caption{Number of SFT samples generated in culinary and tourism domains across three iterative stages. Each stage reflects one cycle of the SearchInstruct pipeline applied to address feedback from the previous round.}
\label{tab:data_turism_culinary}
\end{table}

To evaluate the effectiveness of the proposed method, we conducted a blind human evaluation. Five independent annotators—none of whom were involved in seed construction—were asked to each design 20 diverse questions, resulting in a 100-question benchmark. This benchmark was used to evaluate two versions of the MATINA model: one trained prior to SearchInstruct augmentation, and one trained with our additional SFT data. Annotators were blind to the model identities\footnote{Model identifiers were hidden from annotators during evaluation.}.

As shown in Figure~\ref{fig:human_eval}, models trained with SearchInstruct data performed significantly better on previously weak areas. These results validate the method’s utility in real-world settings and highlight its ability to fill content gaps that standard data generation techniques miss.

This setup also enables iterative refinement: once specific weaknesses in the model’s behavior are identified through human evaluation or domain-specific testing, additional instruction–response pairs can be generated using the SearchInstruct framework to target those areas. As shown in Figure~\ref{fig:refinement_loop}, this creates a continuous improvement loop in which domain-specific shortcomings are systematically addressed. The process can be repeated until performance stabilizes or marginal gains diminish, allowing for efficient and focused model enhancement over time.

\begin{figure*}[t]
    \centering
    \includegraphics[width=0.7\textwidth]{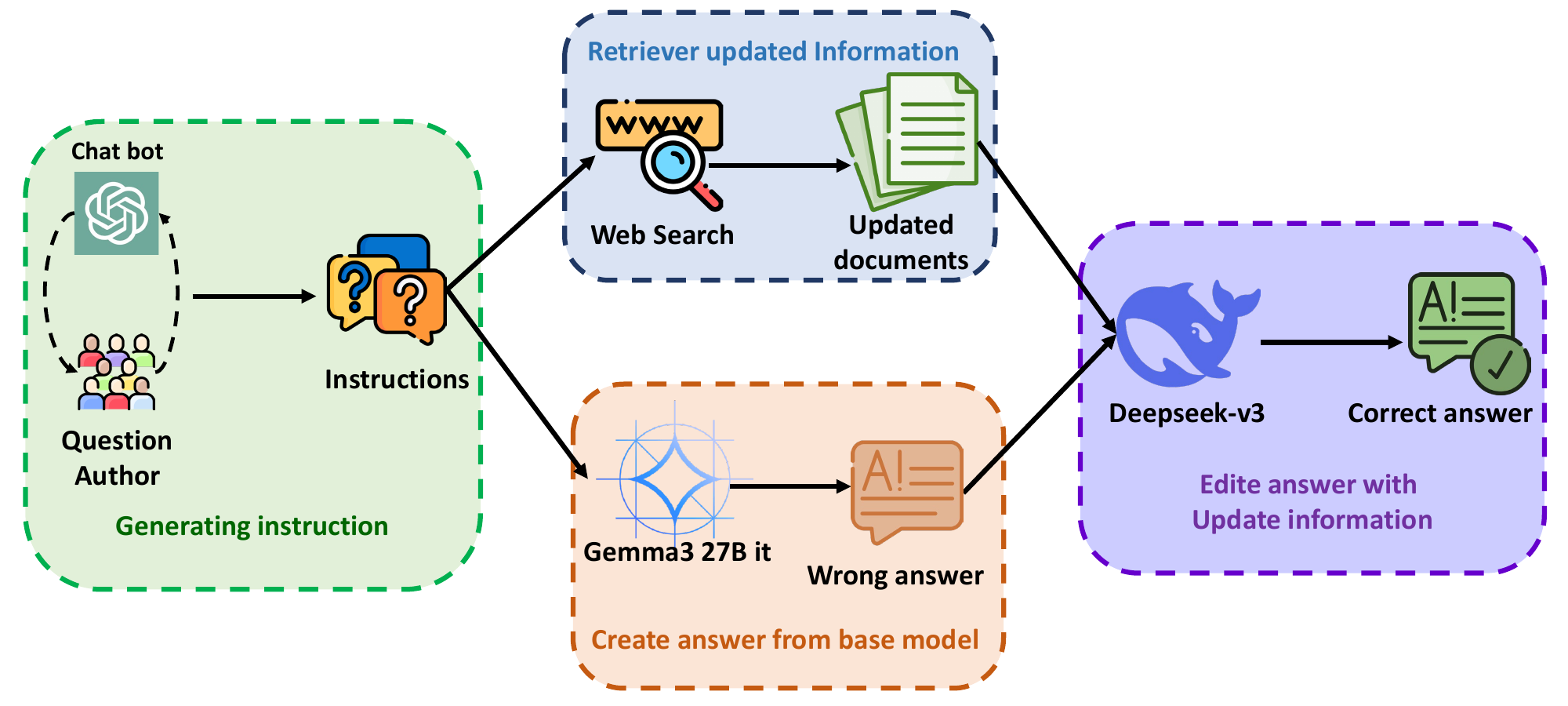}
    \caption{Pipeline for constructing update-specific instruction data used in model editing. Starting from user-provided queries, relevant documents are retrieved and used to construct grounded instruction–answer pairs.}
    \label{fig:model_editing_data}
\end{figure*}

\begin{table*}[ht]
    \centering
    \resizebox{0.6\textwidth}{!}{
    \begin{tabular}{lcc}
        \hline
        \textbf{Category} & \textbf{Gemma3 27B (original)} & \textbf{Gemma3 27B (updated)} \\
        \hline
        STEM            & 66.04 & 62.86 \textcolor{red}{(-3.18)} \\
        Social Sciences & 79.30 & 77.87 \textcolor{red}{(-1.43)} \\
        Humanities      & 60.60 & 59.02 \textcolor{red}{(-1.58)} \\
        Other           & 73.66 & 71.65 \textcolor{red}{(-2.01)} \\
        \rowcolor{orange!10}
        Average         & 68.88 & 66.89 \textcolor{red}{(-1.99)} \\
        \hline
    \end{tabular}
    }
    \caption{Performance of Gemma3 27B before and after the update. Score decreases are highlighted in red.}
    \label{tab:mmlu}
\end{table*}

\subsection{Model Updating via SearchInstruct}

Another practical use case for the SearchInstruct framework is model updating—specifically, targeted modification of model outputs based on recent information. The underlying intuition is that by retrieving up-to-date documents relevant to a query, it is possible to revise the model's response in a lightweight and localized manner.

To constrain the task for evaluation purposes, we focused on a narrow update scope: tracking recent political changes and selected current events. Our goal was to update only specific knowledge while preserving the rest of the model’s internal representations, effectively performing controlled \textit{model editing}.

We applied this strategy to the \texttt{Gemma 27B} model. First, using a set of manually curated seed instructions (see \hyperref[subsec:model-updating]{Appendix C.3}), we generated query–response pairs via SearchInstruct. For each instruction, we obtained the model’s original output and retrieved relevant, up-to-date documents. These documents included new facts—such as political appointments, resignations, or verified current affairs.

To incorporate the updated knowledge, we used a secondary model such as \texttt{DeepSeek} to minimally revise the original answer from Gemma. The editing model was guided to only modify incorrect or outdated information based on the retrieved documents, without altering unrelated content. The prompt system used for editing is described in \hyperref[appendix:answer-updating]{Appendix A.4}.

As a result, only selected spans within the model’s responses were modified, ensuring that the updates were localized, accurate, and minimally invasive. A schematic overview of the update-specific data generation process is shown in Figure~\ref{fig:model_editing_data}, and examples of before-and-after outputs are provided in \hyperref[appendix:updated-examples]{Appendix D}.

\subsubsection{Evaluation of the Edited Model}

To evaluate the effectiveness of model updating via SearchInstruct, we constructed a preference dataset for fine-tuning using the \texttt{ORPO} \cite{hong2024orpo}. For each instruction $i \in I^{\text{edit}}$ (instructions related to recent knowledge), we defined a pair:
\[
(i, A_i^{\text{reject}}, A_i^{\text{chosen}})
\]
where $A_i^{\text{reject}}$ is the original output from the base model (e.g., Gemma), and $A_i^{\text{chosen}}$ is the edited response generated by a secondary editing model (e.g., DeepSeek) based on up-to-date documents.

Using this preference data, we fine-tuned the model with ORPO to align its outputs toward the corrected information. We then evaluated the updated model in two ways:

\begin{itemize}[itemsep=0.1em, topsep=0.3em]
    \item \textbf{Targeted evaluation:} A new set of instructions related to the same domain (e.g., political changes) was submitted to both the original and edited models. Human evaluation confirmed that updated responses were correct and aligned with recent information.
    \item \textbf{General evaluation:} We also measured model performance on the \texttt{MMLU} benchmark ~\cite{hendrycks2021measuring}  to assess whether general knowledge was degraded. Results  in Table~\ref{tab:mmlu} show no significant drop in accuracy, suggesting that the editing process preserved the broader knowledge of the model.
\end{itemize}

However, further analysis revealed a critical limitation. In some cases, the model was only capable of answering specific updated questions correctly (e.g., "Who is the current president of Iran?"), but lacked deeper knowledge about the subject. For instance, although the edited model correctly answered “Masoud Pezeshkian” as the current president, it failed to provide accurate information about his past roles or political background—facts not present in the updated dataset and not previously encoded in the model’s internal knowledge.

This indicates that the editing process effectively replaces surface-level facts for specific queries but does not deeply integrate new knowledge into the model’s reasoning capabilities. While model editing is efficient and minimally invasive, its ability to introduce deeply connected knowledge remains limited.

\section{Conclusion}

In this work, we introduced \textit{SearchInstruct}, a novel framework for constructing high-quality instruction datasets for SFT. The method combines LLM-driven instruction generation with targeted document retrieval to address challenges related to diversity, realism, and recency—especially in specialized domains.

We demonstrated the effectiveness of this approach in two key scenarios: (1) enhancing model performance in cultural domains such as Iranian cuisine and tourism, and (2) updating factual knowledge in large language models using minimally invasive editing strategies. Experimental results showed that SearchInstruct can generate contextually accurate and domain-aligned data, leading to measurable improvements without degrading existing knowledge.

Overall, SearchInstruct provides a flexible and scalable solution for data construction and model editing, making it a promising tool for the continued development of adaptive and domain-aware language models.

\section{Limitations}

Despite its effectiveness, the \textit{SearchInstruct} framework has several limitations:

\begin{itemize}
    \item \textbf{Dependency on retrieved documents:} The quality and accuracy of generated answers are directly tied to the reliability of retrieved content. In domains with sparse or noisy resources, this can limit performance.
    
    \item \textbf{Shallow model editing:} Our findings suggest that editing responses through surface-level substitution does not lead to deep conceptual integration. The model may memorize specific updates without understanding broader context.
    
    \item \textbf{Limited scalability:} Although semi-automated, applying the method across large or fast-changing domains still requires substantial resources for document retrieval, filtering, and validation.

    \item \textbf{Reliance on strong LLMs:} Several stages, including query expansion and response construction, depend on high-quality LLMs. In low-resource settings, performance may degrade.

    \item \textbf{Risk of bias propagation:} Web-based retrieval introduces potential biases from source documents, which can affect the neutrality and fairness of generated data.
\end{itemize}

\bibliography{custom}

\begin{thebibliography}{18}
\providecommand{\natexlab}[1]{#1}

\bibitem[{Chen et~al.(2023)Chen, Chen, Goldstein, Huang, and Zhou}]{chen2023instructzero}
Lichang Chen, Jiuhai Chen, Tom Goldstein, Heng Huang, and Tianyi Zhou. 2023.
\newblock \href {https://doi.org/10.48550/arXiv.2306.03082} {Instructzero: Efficient instruction optimization for black‑box large language models}.
\newblock \emph{arXiv preprint arXiv:2306.03082}.

\bibitem[{Guu et~al.(2020)Guu, Lee, Tung, Pasupat, and Chang}]{guu2020realm}
Kelvin Guu, Kenton Lee, Zora Tung, Panupong Pasupat, and Ming‑Wei Chang. 2020.
\newblock Realm: Retrieval‑augmented language model pre‑training.
\newblock In \emph{Proceedings of the 37th International Conference on Machine Learning (ICML 2020)}, pages 3929--3938. PMLR.

\bibitem[{Hendrycks et~al.(2021)Hendrycks, Burns, Basart, Zou, Mazeika, Song, and Steinhardt}]{hendrycks2021measuring}
Dan Hendrycks, Collin Burns, Steven Basart, Andy Zou, Mantas Mazeika, Dawn Song, and Jacob Steinhardt. 2021.
\newblock \href {https://doi.org/10.48550/arXiv.2009.03300} {Measuring massive multitask language understanding}.
\newblock In \emph{Proceedings of the International Conference on Learning Representations (ICLR 2021)}.
\newblock Also available as arXiv:2009.03300.

\bibitem[{Hong et~al.(2024)Hong, Lee, and Thorne}]{hong2024orpo}
Jiwoo Hong, Noah Lee, and James Thorne. 2024.
\newblock \href {https://doi.org/10.18653/v1/2024.emnlp-main.626} {Orpo: Monolithic preference optimization without reference model}.
\newblock In \emph{Proceedings of the 2024 Conference on Empirical Methods in Natural Language Processing}, pages 11170--11189, Miami, Florida, USA. Association for Computational Linguistics.

\bibitem[{Hosseinbeigi et~al.(2025)Hosseinbeigi, SeifKashani, Seraj, Taherinezhad, Nafisi, Nadi, Barati, Hasani, Amiri, and Masoudi}]{hosseinbeigi2025matina}
Sara~Bourbour Hosseinbeigi, MohammadAli SeifKashani, Javad Seraj, Fatemeh Taherinezhad, Ali Nafisi, Fatemeh Nadi, Iman Barati, Hosein Hasani, Mostafa Amiri, and Mostafa Masoudi. 2025.
\newblock \href {https://aclanthology.org/2025.findings-acl.1074/} {Matina: A culturally-aligned {P}ersian language model using multiple {L}o{RA} experts}.
\newblock In \emph{Findings of the Association for Computational Linguistics: ACL 2025}, pages 20874--20889, Vienna, Austria. Association for Computational Linguistics.

\bibitem[{Izacard et~al.(2022)Izacard, Lewis, Lomeli, Hosseini, Petroni, Schick, Dwivedi‑Yu, Joulin, Riedel, and Grave}]{izacard2022atlas}
Gautier Izacard, Patrick S.~H. Lewis, Maria Lomeli, Lucas Hosseini, Fabio Petroni, Timo Schick, Jane Dwivedi‑Yu, Armand Joulin, Sebastian Riedel, and Edouard Grave. 2022.
\newblock \href {https://doi.org/10.48550/arXiv.2208.03299} {Atlas: Few‑shot learning with retrieval augmented language models}.
\newblock \emph{arXiv preprint arXiv:2208.03299}.

\bibitem[{Lazaridou et~al.(2022)Lazaridou, Gribovskaya, Stokowiec, and Grigorev}]{lazaridou2022internet}
Angeliki Lazaridou, Elena Gribovskaya, Wojciech Stokowiec, and Nikolai Grigorev. 2022.
\newblock \href {https://doi.org/10.18653/v1/2022.acl-long.579} {Internet‑augmented language models through search}.
\newblock In \emph{Proceedings of the 60th Annual Meeting of the Association for Computational Linguistics (Volume 1: Long Papers)}, pages 8460--8478, Dublin, Ireland. Association for Computational Linguistics.

\bibitem[{Lewis et~al.(2020)Lewis, Perez, Piktus, Petroni, Karpukhin, Goyal, Küttler, Lewis, Yih, Rocktäschel, Riedel, and Kiela}]{lewis2020retrieval}
Patrick Lewis, Ethan Perez, Aleksandra Piktus, Fabio Petroni, Vladimir Karpukhin, Naman Goyal, Heinrich Küttler, Mike Lewis, Wen‑tau Yih, Tim Rocktäschel, Sebastian Riedel, and Douwe Kiela. 2020.
\newblock Retrieval‑augmented generation for knowledge‑intensive nlp tasks.
\newblock In \emph{Advances in Neural Information Processing Systems 33 (NeurIPS 2020)}, pages 9459--9474. Curran Associates, Inc.

\bibitem[{Lin et~al.(2024)Lin, Chen, Chen, Shi, Lomeli, James, Rodriguez, Kahn, Szilvasy, Lewis, Zettlemoyer, and Yih}]{lin2023radit}
Xi~Victoria Lin, Xilun Chen, Mingda Chen, Weijia Shi, Maria Lomeli, Rich James, Pedro Rodriguez, Jacob Kahn, Gergely Szilvasy, Mike Lewis, Luke Zettlemoyer, and Scott~Wen‑tau Yih. 2024.
\newblock \href {https://doi.org/10.48550/arXiv.2310.01352} {Ra‑dit: Retrieval‑augmented dual instruction tuning}.
\newblock In \emph{Proceedings of the Eighth International Conference on Learning Representations (ICLR 2024)}. OpenReview.net / ICLR.
\newblock Originally published as arXiv:2310.01352 (Oct 2, 2023).

\bibitem[{Ouyang et~al.(2022)Ouyang, Wu, Jiang, Almeida, Wainwright, Mishkin, Zhang, Agarwal, Slama, Ray, Schulman, Hilton, Kelton, Miller, Simens, Askell, Welinder, Christiano, Leike, and Lowe}]{ouyang2022training}
Long Ouyang, Jeffrey Wu, Xu~Jiang, Diogo Almeida, Carroll~L. Wainwright, Pamela Mishkin, Chong Zhang, Sandhini Agarwal, Katarina Slama, Alex Ray, John Schulman, Jacob Hilton, Fraser Kelton, Luke Miller, Maddie Simens, Amanda Askell, Peter Welinder, Paul~F. Christiano, Jan Leike, and Ryan Lowe. 2022.
\newblock \href {https://doi.org/10.48550/arXiv.2203.02155} {Training language models to follow instructions with human feedback}.
\newblock \emph{arXiv preprint arXiv:2203.02155}.

\bibitem[{Sanh et~al.(2022)Sanh, Webson, Raffel, Bach, Sutawika, Alyafeai, Chaffin, Stiegler, Le~Scao, Raja, Dey, Bari, Xu, Thakker, Sharma~Sharma, Szczechla, Kim, Chhablani, Nayak, Datta, Chang, Jiang, Wang, Manica, Shen, Xin Yong, Pandey, Bawden, Wang, Neeraj, Rozen, Sharma, Santilli, Févry, Fries, Teehan, Bers, Biderman, Gao, Wolf, and Rush}]{sanh2022multitask}
Victor Sanh, Albert Webson, Colin Raffel, Stephen~H. Bach, Lintang Sutawika, Zaid Alyafeai, Antoine Chaffin, Arnaud Stiegler, Teven Le~Scao, Arun Raja, Manan Dey, M.~Saiful Bari, Canwen Xu, Urmish Thakker, Shanya Sharma~Sharma, Eliza Szczechla, Taewoon Kim, Gunjan Chhablani, Nihal V. Nayak, and 22 others. 2022.
\newblock Multitask prompted training enables zero‑shot task generalization.
\newblock In \emph{The Tenth International Conference on Learning Representations (ICLR)}, Virtual Event.

\bibitem[{Taori et~al.(2023)Taori, Gulrajani, Zhang, Dubois, Li, Geng, Narayanan, Liang, and Zhang}]{taori2023alpaca}
Rohan Taori, Ishaan Gulrajani, Tianyi Zhang, Yann Dubois, Xuechen Li, Xinyang Geng, Deepak Narayanan, Percy Liang, and Tatsunori~B. Zhang. 2023.
\newblock Stanford alpaca: An instruction‑following llama model.
\newblock \url{https://github.com/tatsu-lab/stanford_alpaca}.

\bibitem[{Wang et~al.(2024)Wang, Ping, McAfee, Xu, Li, Shoeybi, and Catanzaro}]{wang2023instructretro}
Boxin Wang, Wei Ping, Lawrence McAfee, Peng Xu, Bo~Li, Mohammad Shoeybi, and Bryan Catanzaro. 2024.
\newblock \href {https://doi.org/10.48550/arXiv.2310.07713} {Instructretro: Instruction tuning post retrieval‑augmented pretraining}.
\newblock In \emph{Proceedings of the 41st International Conference on Machine Learning (ICML 2024)}, pages 51255--51272. PMLR.
\newblock Originally available as arXiv:2310.07713 (Oct 11, 2023).

\bibitem[{Wang et~al.(2022{\natexlab{a}})Wang, Kordi, Mishra, Liu, Smith, Khashabi, and Hajishirzi}]{wang2022selfinstruct}
Yizhong Wang, Yeganeh Kordi, Swaroop Mishra, Alisa Liu, Noah~A. Smith, Daniel Khashabi, and Hannaneh Hajishirzi. 2022{\natexlab{a}}.
\newblock \href {https://doi.org/10.48550/arXiv.2212.10560} {Self‑instruct: Aligning language models with self‑generated instructions}.
\newblock \emph{arXiv preprint arXiv:2212.10560}.

\bibitem[{Wang et~al.(2022{\natexlab{b}})Wang, Mishra, Alipoormolabashi, Kordi, Mirzaei, Naik, Ashok, Dhanasekaran, Arunkumar, Stap, Pathak, Karamanolakis, Lai, Purohit, Mondal, Anderson, Kuznia, Doshi, Pal, Patel, Moradshahi, Parmar, Purohit, Varshney, Kaza, Verma, Puri, Karia, Sampat, Reddy, Patro, Dixit, Shen, Baral, Choi, Smith, Hajishirzi, and Khashabi}]{wang2022supernaturalinstructions}
Yizhong Wang, Swaroop Mishra, Pegah Alipoormolabashi, Yeganeh Kordi, Amirreza Mirzaei, Atharva Naik, Arjun Ashok, Arut~Selvan Dhanasekaran, Anjana Arunkumar, David Stap, Eshaan Pathak, Giannis Karamanolakis, Haizhi Lai, Ishan Purohit, Ishani Mondal, Jacob Anderson, Kirby Kuznia, Krima Doshi, Kuntal~Kumar Pal, and 19 others. 2022{\natexlab{b}}.
\newblock Super‑naturalinstructions: Generalization via declarative instructions on 1600+ nlp tasks.
\newblock In \emph{Proceedings of the 2022 Conference on Empirical Methods in Natural Language Processing (EMNLP)}, pages 5085--5109, Abu Dhabi, United Arab Emirates. Association for Computational Linguistics.

\bibitem[{Wei et~al.(2021)Wei, Bosma, Zhao, Guu, Yu, Lester, Du, Dai, and Le}]{wei2021finetuned}
Jason Wei, Maarten Bosma, Vincent Zhao, Kelvin Guu, Adams~Wei Yu, Brian Lester, Nan Du, Andrew~M. Dai, and Quoc~V. Le. 2021.
\newblock Finetuned language models are zero-shot learners.
\newblock In \emph{Advances in Neural Information Processing Systems (NeurIPS)}.

\bibitem[{Xu et~al.(2023)Xu, Sun, Zheng, Geng, Zhao, Feng, Tao, Lin, and Jiang}]{xu2023wizardlm}
Canwen Xu, Qingfeng Sun, Kai Zheng, Xiubo Geng, Pu~Zhao, Jiazhan Feng, Chongyang Tao, Qingwei Lin, and Daxin Jiang. 2023.
\newblock \href {https://doi.org/10.48550/arXiv.2304.12244} {Wizardlm: Empowering large language models to follow complex instructions}.
\newblock \emph{arXiv preprint arXiv:2304.12244}.

\bibitem[{Ziegler et~al.(2024)Ziegler, Perez, Pimentel, Reif, Zhang, Manning, and Raffel}]{ziegler2024craft}
Zachary Ziegler, Ethan Perez, Tiago Pimentel, Emily Reif, Tianyi Zhang, Christopher Manning, and Colin Raffel. 2024.
\newblock Craft: Conceptual retrieval‑augmented fine‑tuning.
\newblock In \emph{Proceedings of the 2024 Conference on Empirical Methods in Natural Language Processing (EMNLP 2024)}. Association for Computational Linguistics.

\end{thebibliography}

\clearpage
\appendix
\section*{Appendix A \\ System Prompt Designs}
\addcontentsline{toc}{section}{Appendix A. System Prompt Designs}

Each stage of our pipeline uses a dedicated system prompt tailored to the specific objective of that component. This appendix presents the design of each prompt through structured illustrations.

\subsection*{A.1 System Prompt for Query Expansion}
\label{subsec:query-expansion}
\begin{figure*}[!t]
    \centering
    \includegraphics[width=\textwidth]{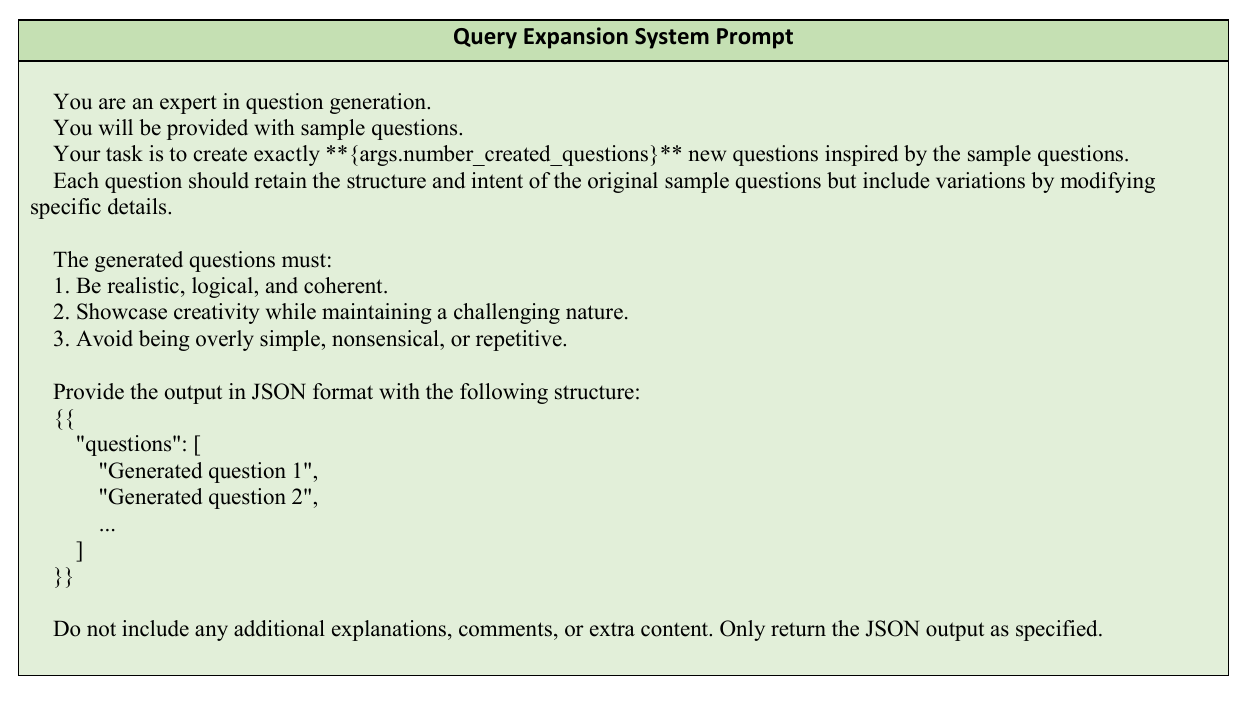}
    \caption{System prompt design for query expansion.}
    \label{fig:prompt_query_expansion}
\end{figure*}

The system prompt used in this stage instructs the model to generate diverse and semantically rich variations of an initial instruction seed. This system prompt is illustrated in Figure \ref{fig:prompt_query_expansion}

\subsection*{A.2 System Prompt for Search-Oriented Rewriting}
\label{appendix:search_prompt}
This prompt transforms general natural language instructions into search-optimized queries that enhance document retrieval performance.

\begin{figure}[H]
    \centering
    \hspace*{-0.2cm}
    \includegraphics[width=0.5\textwidth]{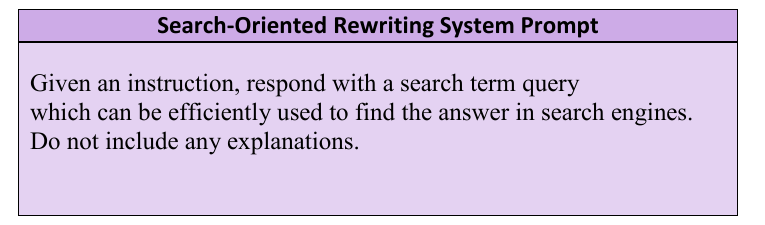}
    \caption{System prompt design for search-oriented query rewriting.}
    \label{fig:prompt_search_rewriting}
\end{figure}

\subsection*{A.3 System Prompt for Response Construction}
\label{appendix:response}

This prompt guides the model to synthesize a grounded and context-aware response using retrieved documents as input.

\begin{figure}[H]
    \centering
    \hspace*{-0.2cm}
    \includegraphics[width=0.5\textwidth]{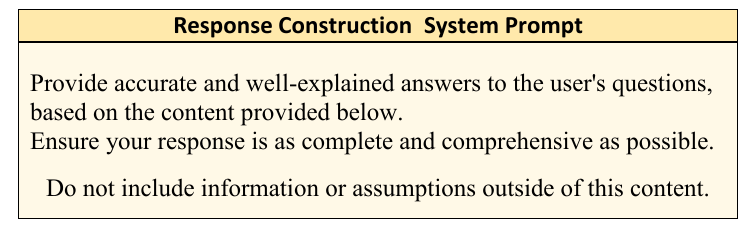}
    \caption{System prompt design for evidence-grounded response construction.}
    \label{fig:prompt_response_construction}
\end{figure}

\subsection*{A.4 System Prompt for Answer Updating}
\label{appendix:answer-updating}
The final system prompt is designed to apply local edits to existing answers when new or corrected information is introduced.

\begin{figure*}[t]
    \centering
    \includegraphics[width=\textwidth]{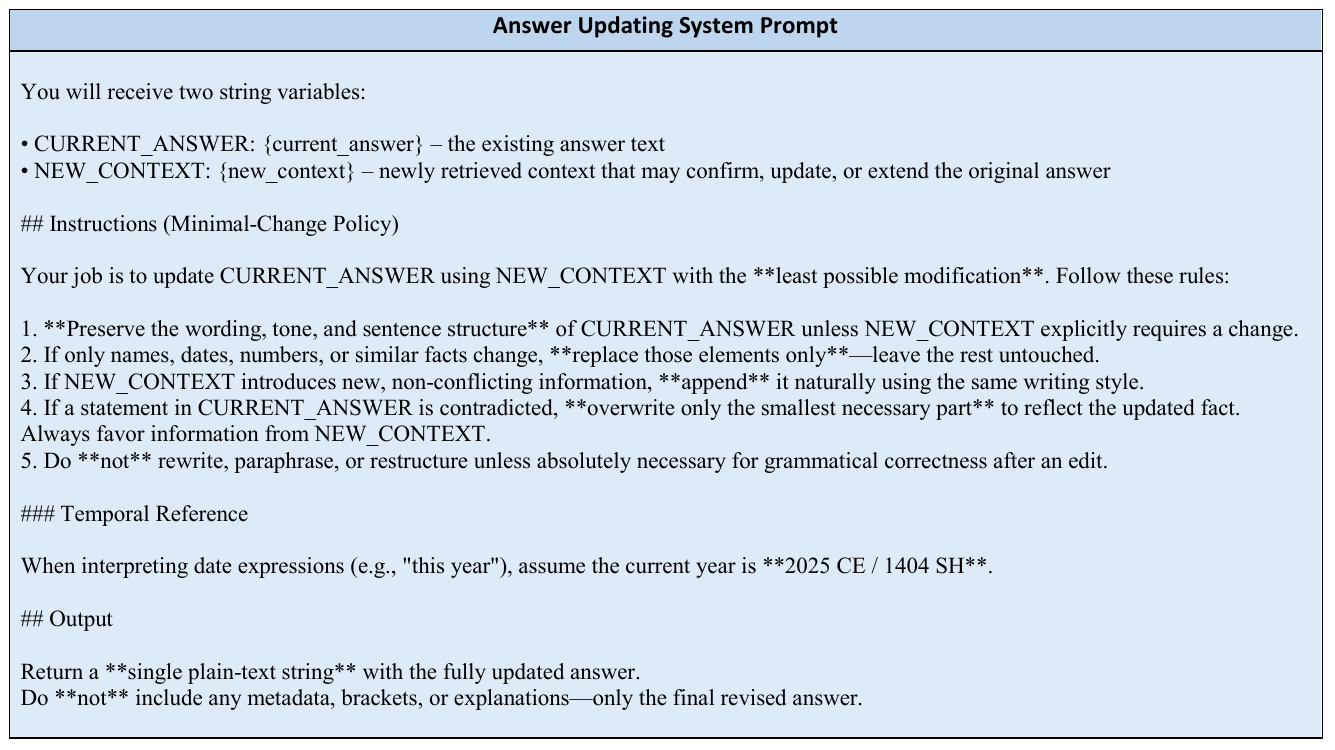}
    \caption{System prompt design for answer refinement and updates.}
    \label{fig:prompt_answer_update}
\end{figure*}

\section*{Appendix B \\ Failure Cases of Current LLM Models}
\label{appendix:b}
\addcontentsline{toc}{section}{Appendix B. Failure Cases of Current LLM Models}

As discussed in the main body of the paper, current language models often fail to provide accurate answers to questions in specialized domains due to insufficient domain knowledge. However, when provided with relevant supporting documents, their performance improves significantly. This observation reflects one of the core ideas behind our proposed method, \textit{SearchInstruct}.

\begin{figure*}[t] 
    \centering
    \includegraphics[width=\textwidth]{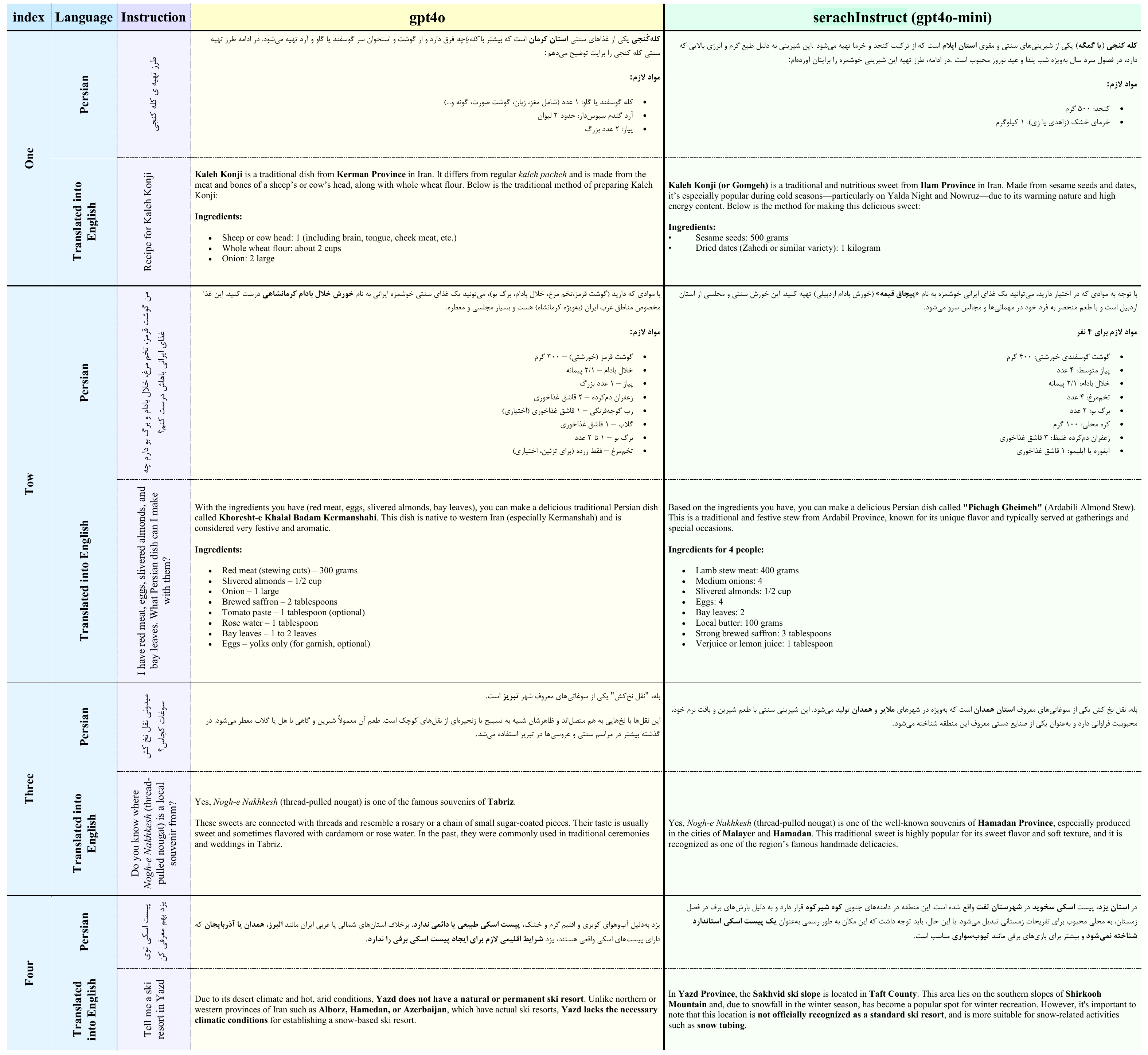}
    \captionsetup{type=table}
    \caption{Examples of partial outputs from  GPT-4o and SearchInstruct (GPT4o mini):This demonstrates that in certain instructions, the SearchInstruct method is capable of producing better responses than larger models, as it benefits from access to highly relevant contextual information, even when using a smaller model}
    \label{tab:gpt4o_vs_searchInstrcut}
\end{figure*}

To demonstrate the effectiveness of our approach in situations where the base model fails to produce a reliable response, we present a set of representative examples in Table~\ref{tab:gpt4o_vs_searchInstrcut}. For each case, we include the model's original response as well as the revised answer generated using the \textit{SearchInstruct} pipeline. These examples highlight how retrieval-augmented prompting can compensate for gaps in model knowledge and substantially improve response quality in specialized domains.

\section*{Appendix C \\ Seeds Generation}
\label{subsec:seed-generation}

To generate high-quality SFT data, we begin with a curated set of instruction seeds $Q$ designed to ensure realism, topic diversity, and domain coverage. We employ two complementary strategies for constructing these seeds:

\textbf{(1) Human-Crafted Seeds.} In the fully manual setup, domain experts were given a detailed instruction-writing guideline that included:

\begin{itemize}
    \item Requirements for seed diversity and contextual realism;
    \item Emphasis on question types that are difficult to synthesize from retrieved documents (e.g., abstract reasoning, subjective judgment);
    \item A comprehensive taxonomy of subcategories within each domain.
\end{itemize}

Each annotator was tasked with writing several domain-specific instructions that reflect real-world user needs and domain-specific reasoning challenges. This ensures that $Q$ contains high-coverage, high-quality questions that may not emerge from automatic generation methods.

\textbf{(2) Human–LLM Collaborative Seeds.} In the hybrid setup, annotators used powerful LLMs (e.g., GPT-4o, Claude 3.5 Sonnet, Gemini 2 Pro) to assist with ideation and rewriting. Annotators were instructed to:

\begin{enumerate}
    \item Ask the LLM to suggest a list of subtopics within the domain;
    \item For each subtopic, prompt the LLM to generate instruction-style questions based on predefined types;
    \item Select, edit, or rewrite the outputs into final seeds that match our quality criteria.
\end{enumerate}

This collaborative setup enabled faster idea generation, controlled diversity, and linguistic variation, while maintaining human oversight to ensure instructional clarity and domain relevance.

In the following sections, we describe the seed construction process for each target domain in detail.

\subsection*{C.1 Culinary Domain}

To construct seeds for the culinary domain, we followed a dual strategy: fully human-written instructions and human–LLM collaborative generation. Our objective was to capture real-world cooking challenges that require reasoning beyond standard recipes, while ensuring topical diversity and realistic phrasing.

\subsubsection*{C.1.1 Human-Crafted Seeds}

In the manual setup, domain experts wrote a wide range of instruction-style questions grounded in common cooking situations. Rather than relying on simple recipe queries, the focus was placed on mistakes, decision-making, ingredient constraints, and reasoning-oriented prompts. Example questions include:

\begin{itemize}
    \item “My Fesenjan stew turned out too sour. How can I fix the taste?”
    \item “How can I grill kebab without using a charcoal grill?”
\end{itemize}

To encourage topic diversity, annotators were provided with a guide listing more than 30 thematic categories. These included examples such as food preservation, regional dishes, dietary limitations, scientific explanations of cooking phenomena, ingredient reuse, children-oriented meals, and presentation aesthetics. These categories served purely as inspiration, not constraints, and annotators were encouraged to propose novel or cross-cutting question types that go beyond the listed themes.

\subsubsection*{C.1.2 Human–LLM Collaborative Seeds}

In the collaborative setup, annotators were encouraged to interact with large language models (e.g., GPT-4o, Claude 3.5) to co-develop seeds. This included using the model to generate candidate questions, rewrite drafts in different tones, or suggest prompts within a topic area.

Common interaction patterns included:
\begin{itemize}
    \item “Suggest a question about common mistakes in cooking rice.”
    \item “Rewrite this to be more natural: Why is the center of my cake undercooked?”
\end{itemize}

Human reviewers selected and refined the outputs to ensure clarity, relevance, and natural phrasing. This collaborative process enhanced topical coverage and linguistic diversity, while accelerating seed creation.

\subsection*{C.2 Tourism Domain}

To construct seeds in the tourism domain, we aimed to capture real-world questions that reflect the complex needs, constraints, and reasoning challenges of travelers within Iran. The objective was to go beyond basic location lookups or generic suggestions and instead focus on experiential, cultural, seasonal, and logistical aspects of domestic tourism.

We adopted a dual strategy: (1) fully human-authored seeds and (2) human–LLM collaborative construction.

\subsubsection*{C.2.1 Human-Crafted Seeds}

In the manual setup, expert annotators were instructed to write instruction-style questions that simulate authentic user concerns in diverse travel scenarios. These included comparisons between destinations, cultural experiences, regional infrastructure challenges, weather-related planning, and cost-aware decisions.

To ensure coverage, annotators were given a guide listing more than 30 example categories. These included: historical landmarks, local cuisines, nature tourism, rural and nomadic travel, eco-tourism, handicrafts and souvenirs, accommodation types, seasonal destinations, transportation, adventure activities, underexplored sites, wellness tourism, and even environmental concerns such as climate impact on desert and coastal areas.

These categories were intended for inspiration only, not as limitations. Annotators were encouraged to mix themes (e.g., cultural + seasonal + budget-focused) and explore niche or unconventional travel scenarios.

Representative examples of well-formed questions include:
\begin{itemize}
    \item “If I want to travel in winter to a warm region with eco-lodges and low tourist density, where should I go?”
    \item “What makes staying in a traditional Yazdi house different from a modern hotel, and which one would be better for a three-day cultural trip?”
    \item “During my visit to Kandovan, the cave-like architecture stood out. How does it compare to stepped villages like Masuleh in terms of tourism experience?”
    \item “Why is Hormuz Island, although less famous than Qeshm or Kish, often described as more unique and immersive?”
\end{itemize}

These questions require the model to reason across multiple dimensions, including cultural insight, regional comparison, travel logistics, and user preference.

\subsubsection*{C.2.2 Human–LLM Collaborative Seeds}

In the collaborative setup, annotators interacted with large language models (e.g., GPT-4o, Claude 3.5) to generate, refine, or extend questions. They could prompt the model with a general topic or theme and request multiple formulations, rewrites, or angle shifts.

Typical prompts used in the process included:
\begin{itemize}
    \item “Generate questions about seasonal travel in northern Iran.”
    \item “Give me a few ideas on how to ask about nomadic tourism experiences.”
\end{itemize}

Annotators reviewed and adjusted the model’s output for realism, clarity, and tone. This approach supported exploration of lesser-covered subdomains, stylistic variation, and faster seed creation, especially in cases where human inspiration plateaued.

Together, these two methods enabled the creation of a rich, diverse, and high-quality set of instruction seeds for the tourism domain, covering both typical and edge-case user needs.

\subsection*{C.3 Model Updating Domain}
\label{subsec:model-updating}

Language models are often unable to respond accurately to questions involving real-world facts that have changed after the model’s training cut-off. These include updates to public figures, international events, institutional changes, or recent sports outcomes. In this section, we focus on designing instruction seeds that explicitly aim to identify and correct outdated model knowledge in such contexts.

The goal was to create seed questions that prompt the model to revise its internal responses, particularly in scenarios where factual updates are essential. These seeds support fine-tuning or model editing workflows and serve as a targeted evaluation tool for temporal generalization.

Seeds were manually constructed by identifying areas where the model is likely to return outdated or incorrect answers. Each question was designed to be direct, fact-seeking, and anchored in well-known public events. Special care was taken to select topics that are newsworthy and important yet unlikely to be politically sensitive in the context of scientific publication.

Representative examples include:

\begin{itemize}
    \item ``Who won the 2024 U.S. presidential election?''
    \item ``Who is currently the president of Iran in 1404 (2025)?''
    \item ``Has the war between Russia and Ukraine ended or is it still ongoing?''
    \item ``What was Iran’s medal ranking in the 2024 Summer Olympics?''
\end{itemize}

Such questions are simple in structure but powerful in purpose: they help surface blind spots in the model’s temporal awareness and provide clear entry points for knowledge injection. The resulting dataset serves both as an editing benchmark and as fine-tuning input for time-sensitive model correction.

\subsection*{Appendix D \\ Examples from the Updated Dataset}
\label{appendix:updated-examples}

This section presents examples from the final dataset constructed for model updating. These instances are designed to help the model revise outdated or incorrect knowledge in a localized and controlled manner, without disrupting its broader behavior.

Each entry consists of a time-sensitive question accompanied by two responses:

\begin{itemize}
    \item \textbf{Rejected:} An outdated, incorrect, or uncertain response—typically resembling what an unedited, pre-trained model might produce.
    \item \textbf{Chosen:} A minimally edited and factually updated version of the rejected answer.
\end{itemize}

\textbf{A key principle in constructing the chosen responses is that the edits are minimal and highly targeted.} Instead of rewriting the entire answer, only the incorrect portion is replaced with updated information, while preserving the original structure, phrasing, and tone as much as possible. This design allows the model to learn localized corrections through fine-tuning or model editing, rather than general retraining.

\begin{figure*}[!t] 
    \centering
    \includegraphics[width=\textwidth]{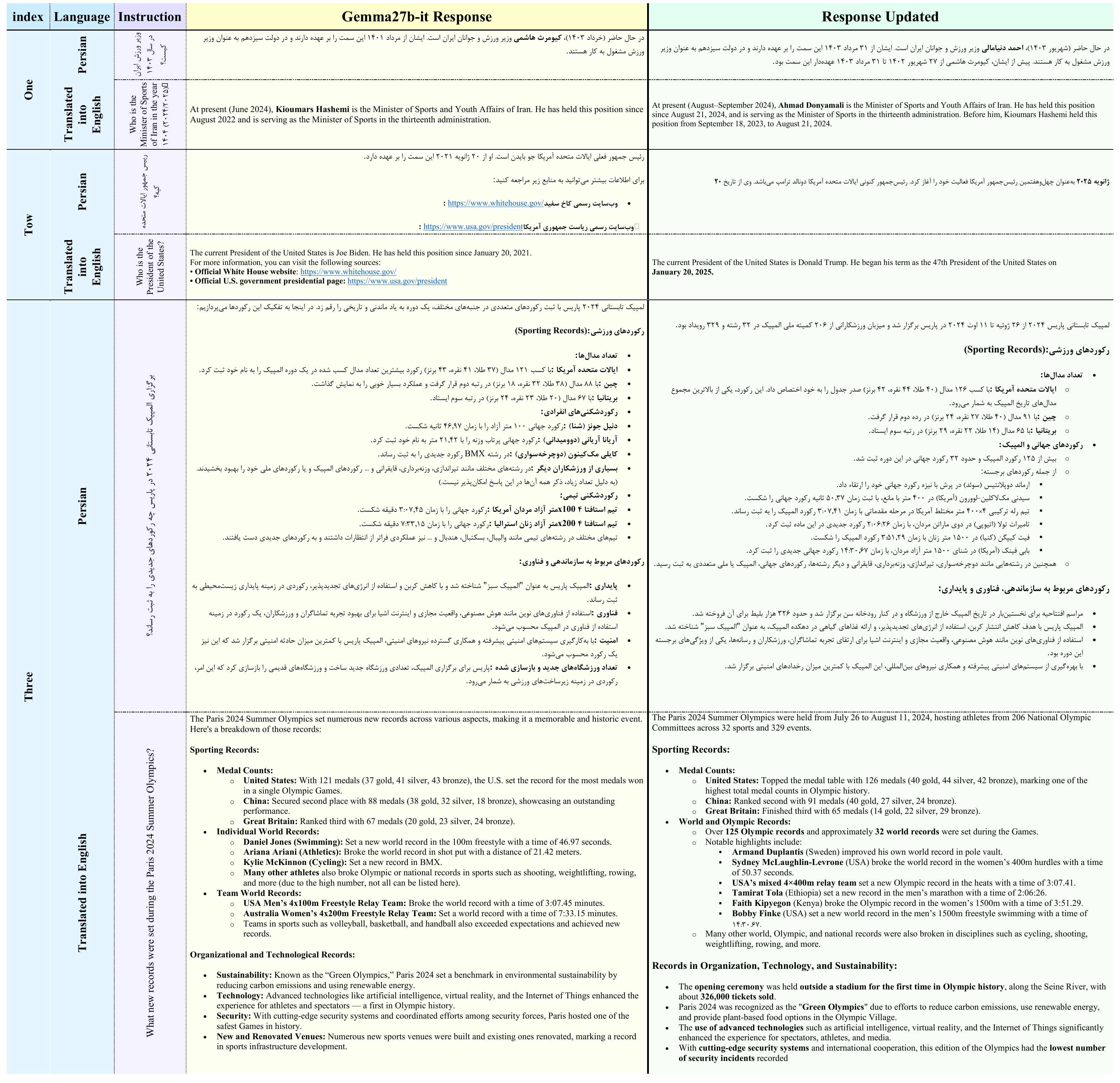}
    \captionsetup{type=table}
    \caption{Representative examples of minimally revised responses with factual updates. Each row includes a user instruction, a rejected response containing outdated or incorrect information, and a chosen response with concise, targeted edits. The examples cover domains such as politics, international affairs, and sports, demonstrating factual correction with minimal changes to phrasing and tone.}
    \label{tab:updated_examples}
\end{figure*}

Table~\ref{tab:updated_examples} provides several representative samples. Each row shows a real-world question, the outdated (rejected) model response, and the minimally revised, accurate version (chosen). These examples span topics such as elections, leadership updates, international conflicts, and sports results.

\section*{Appendix E – Training Hyperparameters}

This appendix provides a consolidated overview of the most relevant training hyperparameters used across different stages of our experiments. We categorize the configurations into three groups:

\begin{itemize}
    \item \textbf{SFT (Supervised Fine-Tuning)} on domain-specific tasks using the Matina-8B and Matina-70B models, applied separately to culinary and tourism instruction datasets.
    \item \textbf{ORPO (Optimal Response Preference Optimization)} training on the \texttt{Gemma-3-27B-IT} model using LoRA adapters and a custom ORPO objective.
\end{itemize}

Table~\ref{tab:hyperparams_unified} presents a unified summary of these hyperparameters, including key factors such as LoRA configuration, learning rate, dropout rate, batch size, training duration, and preference optimization parameters (where applicable)

\begin{table*}[t]
\centering
\small
\begin{tabular}{@{}lcccccccc@{}}
\toprule
\textbf{Model / Task} & \textbf{LoRA Rank} & \textbf{LoRA Alpha} & \textbf{DeepSpeed Stage} & \textbf{Learning Rate} & \textbf{Dropout} & \textbf{Batch Size} & \textbf{Epochs} & \textbf{Pref.\,$\beta$} \\
\midrule
\multicolumn{9}{l}{\textit{Matina-8B (SFT)}} \\[2pt]
Culinary & 128 & 256 & Z0 & $1\times10^{-4}$ & 0.05 & 64 & 3 & -- \\
Tourism  & 256 & 512 & Z0 & $1\times10^{-4}$ & 0.05 & 64 & 4 & -- \\
\cmidrule(lr){1-9}
\multicolumn{9}{l}{\textit{Matina-70B (SFT)}} \\[2pt]
Culinary & 128 & 256 & Z3 & $1\times10^{-4}$ & 0.05 & 8 & 2 & -- \\
Tourism  & 256 & 512 & Z3 & $1\times10^{-4}$ & 0.05 & 8 & 3 & -- \\
\cmidrule(lr){1-9}
\multicolumn{9}{l}{\textit{Gemma-3-27B-IT (ORPO)}} \\[2pt]
Update  & 16  & 32  & Z3 & $2\times10^{-5}$ & 0.005 & 4 & 6 & 0.05 \\
\bottomrule
\end{tabular}
\caption{Unified view of key hyperparameters for all training stages. 
Rows are grouped into three sections: SFT with Matina-8B, SFT with Matina-70B, and ORPO on Gemma-3-27B-IT.}
\label{tab:hyperparams_unified}
\end{table*}

\section*{Appendix F: Future Work}
\label{appendix:future_work}

While \textit{SearchInstruct} demonstrates promising results, several directions remain open for future exploration:

\begin{itemize}
    \item \textbf{Deeper exploration of RAG methods and external APIs:}  
    Our current pipeline only briefly references the use of retrieval-augmented generation (RAG) techniques and external APIs, without a thorough analysis of their practical impact. Future work could investigate how different RAG configurations, API sources, and retrieval strategies affect data diversity and factual accuracy across domains.

    \item \textbf{Leveraging agent-based multi-step interactions for RL training:}  
    Large language model agents are increasingly used to perform multi-step tasks such as guided retrieval, source evaluation, answer synthesis, and refinement. Capturing these rich, interactive sequences offers an opportunity to create environments or trajectories for reinforcement learning (RL). This could facilitate training of more capable agents in real-world, information-seeking scenarios.

    \item \textbf{Integration with structured knowledge sources:}  
    Incorporating structured resources—such as knowledge graphs, ontologies, or semantic lexicons—could improve answer precision and coherence, especially in knowledge-intensive domains like law, medicine, or education. Future pipelines may combine retrieved text with structured representations to enhance factual grounding.

    \item \textbf{Automated optimization of the feedback loop:}  
    Our current framework includes an iterative improvement loop driven by human feedback and manual adjustments. A promising future direction is to formalize this loop using reinforcement learning or other optimization techniques, allowing the system to continuously refine its query generation, retrieval, and synthesis strategies based on performance signals.

    \item \textbf{Automated quality assessment tools:}  
    Evaluating the quality of generated training data remains a bottleneck. Future work could explore automatic metrics or learned models that assess dimensions such as evidential coverage, factual accuracy, and stylistic consistency, reducing reliance on manual review.
\end{itemize}

\end{document}